# Structure leads and dominates comprehension in naturalistic reading

Nan Wang[1, 2], Hanlin Wu[2], Jiaxuan Li[3]

1 Department of Brain and Cognitive Sciences, University of Rochester

2 Department of Linguistics and Modern Languages, the Chinese University of Hong Kong;

3 Department of Language Science, University of California Irvine;

Corresponding author: Nan Wang, nwang40@ur.rochester.edu

Abstract: The hierarchical account and statistical or sequential account have long been framed as rival theories in explaining online comprehension. A lot of evidence has shown that both hierarchical and non-hierarchical factors can shape comprehension and the open question is no longer whether hierarchy contributes, but when and how strongly it does. We addressed the question with co-registered EEG and eye-tracking, treating syntactic depth as the variable for operationalizing hierarchical structure. On timing, hierarchical structure influenced reading before the eyes fixated a word: its neural effect emerged as early as 108 ms before fixation onset, over right-central regions, and the scanpath showed an anticipatory bias toward structurally central words. Both the transitional-probability analysis and the regression on fixation-related potentials supported this pre-fixational timing. In the transitional-probability analysis, readers preferentially moved between syntactically central words rather than following serial word order, showing that scanpaths are organized by syntactic depth rather than by linear adjacency. On strength, Bayesian network modeling showed that syntactic depth was the strongest predictor of departures from linear, word-by-word reading, outweighing lexical familiarity and surprisal. Taken together, the results indicate that hierarchical structure anticipatorily guides online comprehension at both the behavioral and neural levels, and dominates the reading path relative to statistical features.

# 1. Introduction

In naturalistic reading or listening, our comprehension system quickly constructs an interpretation of what was conveyed. Decades of work indicate that two kinds of information drive this process. The hierarchical account assumed that comprehension is driven by the construction of phrase structure and syntactic dependencies that are not reducible to linear word order or statistics (Frazier & Rayner, 1982; Jurafsky, 1996; Vosse & Kempen, 2000). The other side argued that many phenomena can be captured by local statistics or prediction over adjacent words, without strong sensitivity to hierarchical syntactic organization (Christiansen & Chater, 2008; Frank et al., 2012; Frank & Bod, 2011a). Each account has been evaluated primarily against the phenomena it was designed to explain. We review the evidence for each and argue that this separation has impeded resolution of the debate. Further progress requires a direct comparison of the two accounts. The question isn't whether each contributes, but which contributes more strongly and which operates earlier.

## 1.1 Hierarchical account

The central effort of the hierarchical account has been to show that abstract structure exists and that comprehension builds representations not reducible to word order or statistics. This effort draws converging behavioral and neural evidence from four classes of phenomena: online tracking of phrase- and sentence-level structure during connected speech; composition of constituents into combined meanings; formation of long-distance dependencies that override linear distance; and resolution of structural ambiguity. Each class isolates a different way structure makes itself visible, and together they define the standard the sequence-based account must meet.

The first is constituent structure tracking in continuous speech. Ding et al. (2016) firstly showed that cortical activity in MEG concurrently tracks phrase- and sentence-level constituents at distinct frequencies when listening to Mandarin. The dissociation from acoustic cues and raw word predictability offered robust evidence of abstract constituent structure in comprehension and has triggered many similar paradigms (Liu et al., 2026; Zou et al., 2026). ECoG work extended this to localized brain regions, where researchers reported neural sequences in left inferior frontal gyrus that match the composition operations specified by a hierarchical parser (Nelson et al., 2017). The fMRI domain also showed converging picture. Hierarchical structures were translated into parser-derived complexity metrics (node count from context-free and minimalist grammars) and BOLD signal was regressed against these metrics to track abstract structural metrics that go beyond statistical predictability (Brennan et al., 2016).

If the first class shows that structure is tracked, the second shows that it is built. This phenomenon is the composition of constituents into combined meanings in minimal phrases, such as adjective–noun phrases (Bemis & Pylkkänen, 2011; Price et al., 2015). By varying semantic content without introducing syntactic changes, researchers aimed to isolate the brain areas for conceptual and syntactic composition, which were localized to be Left Anterior Temporal Lobe (Westerlund & Pylkkänen, 2014; Zhang & Pylkkänen, 2015), and Angular Gyrus and Left Inferior Frontal Gyrus separately (Hultén et al., 2019; Leiken et al., 2015; Stromswold et al., 1996; Williams et al., 2017).

Thirdly, it's long-distance dependency. The classical behavioral signature of hierarchical dependency formation is the filled-gap effect. Reading times would slow at the position where an expected gap is filled by an overt noun phrase, indicating that comprehenders posit gaps several constituents downstream of a filler (Stowe, 1985). The agreement-attraction literature provides converging evidence: subject–verb agreement is influenced by structurally inaccessible attractor nouns in ways that track hierarchical c-command relations rather than linear adjacency (Dillon et al., 2013; Wagers et al., 2009). Electrophysiologically, sustained left anterior negativities at the position of an expected gap have been reported in Event-related potential (ERP) studies of filler-gap dependencies (Phillips et al., 2005a), and follow-up MEG and fMRI work has localized active dependency formation to left anterior temporal and inferior frontal regions.

Lastly, it's ambiguity resolution and reanalysis. Frazier and Rayner (1982) interpreted these costs as evidence of structure-driven Late-Closure and Minimal-Attachment heuristics, and the broad pattern has been reproduced across many syntactic configurations. ERP work identifies the P600 component as a neural signature of structural reanalysis, with amplitude scaling with the difficulty of the required revision (Kaan et al., 2000a; Osterhout & Holcomb, 1993).

Across these four phenomena, the hierarchical account has built a strong case that abstract structure exists. The case has limits, though. The evidence comes mostly from controlled paradigms and it asks whether structure is present, not whether structure organizes processing as it unfolds. Showing that comprehenders build hierarchy is not the same as showing that hierarchy guides what they do next: where they look, what they hold, what they integrate. Whether hierarchical depth shapes online reading, and whether it does so more strongly than the statistical cues of the competing account, remains untested in natural reading.

### 1.2 Non-hierarchical account

The competing account has been tested against a similar circumscribed set of phenomena it can explain without structure. It comes in two variants, statistical and sequence-based account.

The statistical variant makes a claim about predictors. It holds that surprisal captures much of the variance in reading times and language-selective neural responses, with no need for hierarchical factors (Hale, 2001; Levy, 2008). The account manifests itself across two phenomena. For reading times, the predictive power of word surprisal scales with language-model quality (Goodkind & Bicknell, 2018), and surprisal robustly predicts reading time across large datasets, with a logarithmic link between the two (Shain et al., 2024). For neural responses, researchers often derived the surprisal-related metrics from large transformer models and used them to predict fMRI and ECoG responses (Schrimpf et al., 2021), and even drive and suppress responses in the language network (Tuckute et al., 2021). Comparatively, the statistical account has been much less productive in predicting the magnitude of the disambiguation cost in classical garden-path contexts (Huang et al., 2024; van Schijndel & Linzen, 2019).

Sequence-based account is distinct from statistical accounts in that they take a stronger position about representation. On this view, the moment-to-moment representation is a linear sequence of recent material, perhaps compressed into shallow chunks, but never a hierarchical tree. Frank and Bod (2011b) trained probabilistic language models with either a hierarchical or a sequential (recurrent) architecture to predict reading times in the Dundee corpus; hierarchical models added no variance beyond sequential ones, which led them to conclude that the sentence-processing system shows little online sensitivity to hierarchical structure. Frank et al. (2015) extended this to ERPs, showing that surprisal from sequential architectures predicts N400 amplitude during reading. A memory argument explains why. Because perceptual memory decays within tens to hundreds of milliseconds, comprehension must chunk incoming input into successively higher-level units, and those units form a history of processing operations rather than a stable tree (Christiansen & Chater, 2008). What looks like hierarchy, on this view, is the residue of repeated chunk-and-pass operations on surface sequences.

Like the hierarchical account, then, the non-hierarchical account succeeds within the phenomena built to favor it. Neither side has been made to compete with the other on the same data.

### 1.3 The present study

The debate has remained unresolved because each account has been examined primarily within the phenomena that favor it. More evidence is showing that hierarchical and statistical information make partially distinct and complementary contributions (Brennan et al., 2016; Caucheteux et al., 2023; Gwilliams et al., 2025). Casting results as support-or-refutation of binary oppositions rarely settles anything (Newell, 1973). Settling the debate requires evaluating both accounts jointly, across behavioral and neural data, to determine which factor bears the explanatory load and on which timescale.

The present study targets exactly the gaps identified above: we covered both behavioral and neural data; we evaluated joint role of hierarchical structure and statistical information; we examined both strength and timing of hierarchical structure (Figure 1). Across the analysis, syntactic depth is used as proxy for hierarchical structure processing. It is measured as the number of dependency links traversed from a word to the root of its sentence's parse, and has been used as a proxy for hierarchical complexity in prior neural and behavioral work (Nelson et al., 2017; Weissbart & Martin, 2024).

We pursued this through three analyses on co-registered EEG and eye-tracking from the ZuCo corpus during naturalistic sentence reading. Each analysis is built around a direct contrast between hierarchical feature with the non-hierarchical alternative.

The first analysis contrast hierarchical account with the sequence-based account. The sequence-based view predicts that readers' eye movement should be organized around linearly adjacent material, so scanpaths should track the linear order of the text. We tested this by classifying words in each sentence as syntactic heads (parsing depth 0 or 1) or non-heads (depth $\geq 2$) and computing the four transition probabilities (head-to-head, head-to-non-head, non-head-to-head, non-head-to-non-head) for two sequences: the original left-to-right text order, which embodies the sequence-based prediction of strict linear adjacency, and the fixation-derived sequence, which embodies the actual scanpath. A higher head-to-head transition rate in the fixation-derived sequence relative to the linear baseline, is the signature of a hierarchical syntactic structure guiding reading. It also speaks to the timing of the structure: a bias toward structurally central targets is a pre-fixation, anticipatory signature of hierarchical processing.

The second analysis contrast hierarchical account with the statistical account to further establish the strength of syntactic depth. We measured the magnitude of departures of the fixation-derived sequence from linear text sequence using the edit distance, and included it as the outcome variable in a Gaussian Bayesian network alongside two structural predictors (maximum parsing depth, number of clauses) and two lexical predictors drawn from the E-Z Reader model that operationalize the statistical alternative (sentence familiarity from frequency norms; sentence surprisal from GPT-2) (Radford et al., 2019; Reichle et al., 2003). Network structures were learned and relative weight of each predictor on edit distance is read off the local regression at the outcome node as a measure of strength.

The third analysis moved to the neural domain and set syntactic depth directly against the statistical account at the neural level and established timing of syntactic processing. Fixation-related potentials were extracted and regressed on five standardized predictors: syntactic depth, syntactic surprisal, word length, lexical frequency, and lexical surprisal. Group-level inference on the subject-level coefficient time courses was performed by one-sample cluster-based permutation tests against zero, separately for six regions of interest. The significant time windows give the timing of each predictor in the neural signal.

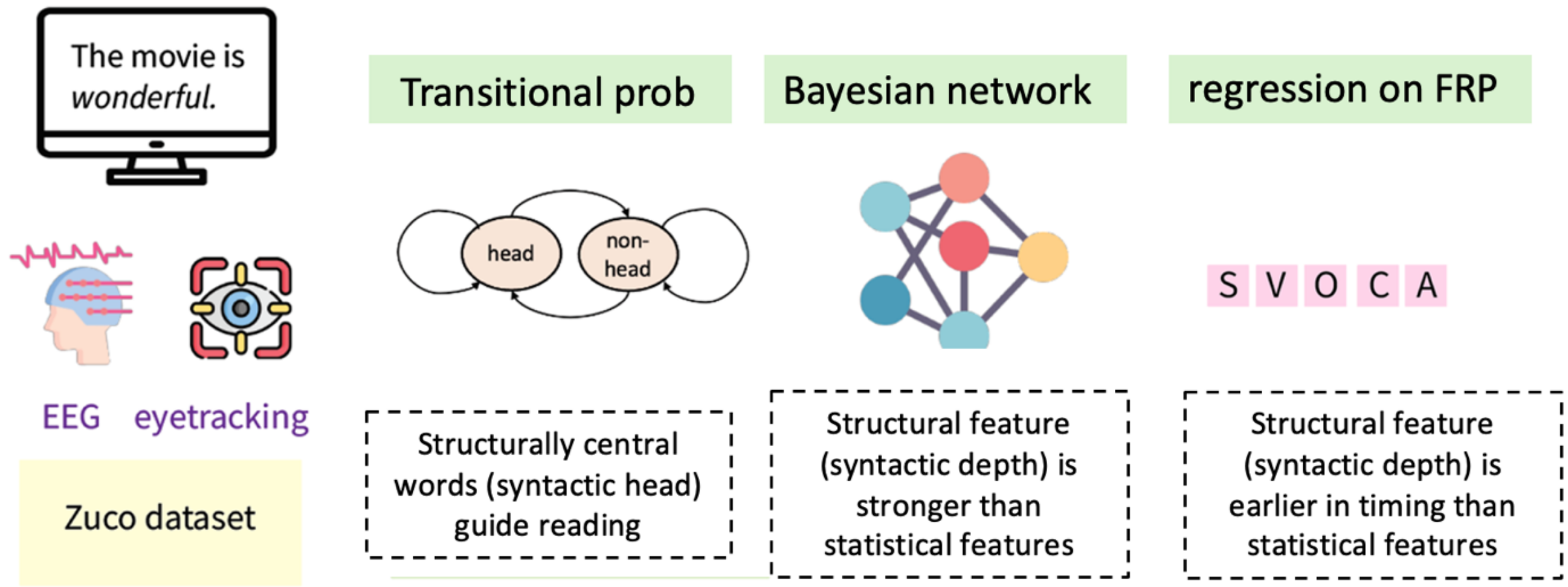

Figure 1. Overview of the methodological pipeline and research aims. Using the ZuCo dataset with co-registered EEG and eye-tracking during naturalistic reading, we implemented three complementary analyses: (i) transitional probability analysis of head and non-head fixations to test the influence of syntactic depth, (ii) Bayesian network modeling to assess the relative strength of syntactic depth compared to other predictors, and (iii) regression on fixation-related potential to examine the timing of syntactic depth in neural response.

## 2. Methods

### 2.1 Materials and experiment design

We used an openly available dataset, Zurich Cognitive Language Processing Corpus, for analysis (Hollenstein et al., 2018). It comprises simultaneous eye-tracking and EEG recordings of 12 native English participants (5 female; mean age = 37.5 years; SD = 10.3 years) during reading. In this task, participants were presented with 400 positive, negative or neutral sentences and rated the emotion valence after reading each sentence. The sentences are movie reviews extracted from Stanford Sentiment Treebank (Socher et al., 2013). A whole sentence was presented each time and participants were allowed to read at their own pace, which is closer to a natural reading scenario, in comparison to rapid serial visual representation. Subjects needed to undergo three reading tasks in total: in the first task, they were presented sentences with positive, negative and neutral emotion valence; in the second task, they were presented sentences with certain semantic relations; in the third task, sentences were similar to those in the second task but participants were instructed to focus on certain relation types during reading. We used data from the first task rather than from the second or third task for three reasons: firstly, the emotional valence task is more likely to elicit holistic sentence processing compared to the second task which involves factual questions and might bias participants to focus on semantic content rather than syntactic structure; secondly, the stimuli in the first task, which are from movie reviews, tend to use more diverse syntactic constructions than the biographical Wikipedia sentences in the second task; thirdly, the omission rates on sentence level are lowest across the three tasks. Eye-tracking data and EEG data were recorded simultaneously during their reading.

### 2.2 Data acquisition and preprocessing

Eye movements were recorded with EyeLink 1000 Plus infrared video-based eye tracker (SR Research) at a sampling rate of 500 Hz, while EEG data were recorded with a 128-channel EEG Geodesic Hydrocel system (Electrical Geodesics, Eugene, Oregon), at a sampling rate of 500 Hz with a bandpass of 0.1 to 100 Hz. The preprocessing of EEG and eye movements followed the same pipeline of Hollenstein (2018). For the eye movements, fixations were identified with a Gaussian mixture model for better cluster fixation data and fixation allocation, and those shorter than 100ms were excluded from further analysis as they are unlikely to reflect meaningful cognitive activities. EEG data underwent high-pass filtering at 0.5Hz and notch filtering at 49-51Hz. Channels with less than 0.85 correlation with other channels, with line noise 4 standard deviations more than other channels, with a flatline duration longer than 5 seconds were marked as bad. Artifacts were removed by regressing out the nine EOG (electrooculography) channels from the 105 scalp EEG channels and MARA (Multiple Artifact Rejection Algorithm) (Winkler et al., 2011). Bad channels that could not be cleaned were interpolated using spherical spline interpolation.

### 2.3 Analysis 1: Transitional probability analysis

The first analysis speaks to the timing of hierarchical structure processing. Because the next fixation target is selected before that word is fixated, any systematic bias in the scanpath toward syntactically central words may suggest that reading behavior is possibly structure-guided. The analysis placed hierarchical account directly against the sequence-based alternative. Sequence-based accounts predict that the reader's moment-to-moment representation is organized around linearly adjacent material, so the fixation transitions should track the linear order of the text. A hierarchical organization predicts the opposite: transitions should preferentially connect words that are central in the dependency tree, including pairs that are non-adjacent on the surface and could not be linked by a representation that is sensitive only to linear adjacency.

We calculated transitional probability between fixated words derived from eye movement data. We focused on the movement of eyes instead of durations of fixations as syntactic analysis related strategies are often not visible in duration-only measures but in scan paths (Dempsey et al., 2023; Malsburg & Vasishth, 2013). Each word in each sentence was assigned a depth value derived from its position in a dependency parse. Parsing depth values were obtained using the dependency module of the Stanford Parser (de Marneffe et al., 2006; Nivre et al., 2016; Schuster & Manning, 2016), which constructs a syntactic tree under the Universal Dependencies formalism. The root of a sentence was considered the base node and was assigned zero and the depths of the rest words were determined by counting the number of dependency links traversed from the word to the root. For example, in the sentence, "The girl lived in the house", the root ("lived") is at the top level and is assigned a depth of zero, while the subject ("girl") and prepositional phrase ("in the house") are directly dependent on the root (see Figure 2 (a)).

These words were categorized into two groups: head (depth = 0 or 1, corresponding to syntactic heads) and non-head (depth ≥ 2). Those with a parsing depth of 1 or 0 were defined as head words as they are most syntactically close to the root, thus playing a prominent role in the sentence. Conversely, words with a higher depth (i.e., those farther from the root, such as modifiers or adjuncts) often serve secondary syntactic roles, and were labeled as non-head. We defined syntactic heads in the strictest sense, and two levels of depths (0 and 1) are the minimum necessary for calculating head-to-head transitional probabilities within a sentence. Including words at depth ≥ 2 as heads would possibly dilute the focus on the core syntactic elements, the distribution of depth values can be seen in Figure 2c.

After labeling each word in each sentence as "head" or "non-head", we computed transitional probabilities for the four possible transition types: head-to-head, head-to-non-head, non-head-to-head, and non-head-to-non-head. The transitional probability of each type is given by dividing the frequency of that transition type by the total number of transitions. We compared the transitional probabilities between two types of sequence. The first sequence followed the linear order of words in the original sentence (e.g., "The girl lived in the house"). The second sequence was fixation-driven: during reading, the eyes do not always move strictly from left to right, but may skip ahead or regress to earlier words, producing a sequence determined by the order of fixated words (e.g. "The girl lived girl in house the house"). The transitional probability derived from the first sequence will form a baseline that reflects probabilities expected under the assumption that readers read sentences without any selective bias towards structurally central words. It will be compared against transitional probability calculated from the second condition.

The central prediction was that, if the scanpath is organized by hierarchical structure rather than by linear adjacency, head-to-head transitions should be elevated in the fixation-derived sequence relative to the linear-text baseline. We tested this with an independent-samples t-test comparing head-to-head transition rates across the two conditions. The remaining three transition types were tested in the same way to assess whether any effect was specific to head-to-head transitions.

The comparison was between conditions (linear-text vs. fixation-derived) within a transition type, rather than between transition types within a single condition. This was a deliberate choice: some sentences may, by virtue of their construction, already contain a high proportion of head-to-head transitions in their linear order, which would inflate the head-to-head probability in the fixation-derived sequence without that elevation reflecting any syntactically guided eye-movement strategy. Comparing across conditions controls for this baseline confound.

In addition, we considered the possibility that the observed effect could be driven by regressions, as regressions often involve returning to earlier words and may artificially increase certain transition types in the fixation-derived sequence. To rule out this possibility, we removed words corresponding to regressions in the fixation-derived sequence and recalculated the transition probabilities for all transition types. If the head-to-head increase remained after removing regression-related words, this suggests that the effect is not merely a byproduct of backward eye movements, but reflects a tendency for readers' fixation sequences to connect syntactic heads.

## 2.4 Analysis 2: Bayesian network analysis

Analysis 2 aimed to quantify how strongly hierarchical structure predicts departures from linear, word-by-word reading and we pitted hierarchical account against statistics-based account in this setting. Structural predictors derived from the dependency-parsed sentence index the hierarchical account, while lexical predictors drawn from the E-Z Reader model index the statistical account.

We used Bayesian-network framework to estimate the joint distribution over variables and to obtain the dependency structure among predictors (via structure learning) and the relative weight of each direct parent of the outcome (via the local linear-Gaussian regression at the outcome node).

*2.4.1 Define variables*

To quantify the extent to which reading departs from serial word order, we used the edit distance between the linear-text sequence and the fixation-derived sequence introduced in Analysis 1. Edit distance is a standard measure of surface-form similarity between two sequences and represents the minimal number of edit operations required to transform one into the other; it was computed using the Needleman–Wunsch algorithm (Needleman & Wunsch, 1970). Larger edit distance indicates a fixation-derived sequence that diverges more from strict left-to-right reading order.

Four other variables were also included. Two were structural variables derived from dependency parsing trees: maximum parsing depth, indexing the depth of the most deeply nested dependency relation in a sentence (Weissbart & Martin, 2024), and the number of clauses, which reflects sentence-level structural complexity (Lu, 2010). The two variables could encode the extent to which the extent to which the reader was tracking a syntactic structure rather than pure statistics.

Two more variables, familiarity and surprisal, were drawn from the E-Z Reader model of eye movement (Reichle et al., 1998, 2003). Sentence familiarity was calculated by averaging lexical-level estimates of word retrieval ease based on frequency norms across sentences. Sentence surprisal was computed using the GPT-2 language model (Radford et al., 2019), as the negative log probability of each word given its preceding context, then averaged across words to obtain a sentence-level measure. These variables were included because they represent core lexical influences on eye movements: familiarity is associated with early lexical access, whereas surprisal captures contextual expectancy (Reichle et al., 2003). Two variables showed strong right skew on the raw scale and were log1p-transformed prior to analysis (edit distance and sentence surprisal); the remaining three were retained on their original scale. Post-transform skewness fell within an acceptable range for Gaussian modeling for all variables, and bivariate scatterplots showed approximately linear relationships, motivating the modeling choices below (See Supplementary S1).

*2.4.2 Network specification, structure learning and parameter fitting*

A Gaussian Bayesian network was estimated with BIC scoring (bnlearn, version 5.0.1, Scutari, 2010). Under this specification, the joint distribution over the five variables factorizes along the directed acyclic graph (DAG), and the local model at each node is a linear-Gaussian regression of that variable on its parents in the DAG. Variables with no parents in the DAG (i.e., the root node) were modeled as marginal Gaussians. The edges flowing from edit distance to other variables were blacklisted on theoretical grounds.

Structure learning of the network used hill-climbing search (HC) as the primary algorithm and it's cross validated with two other algorithms: tabu search and the Max-Min Hill-Climbing algorithm (MMHC) (Glover & Marti, 2006; Heckerman et al., 1995; Tsamardinos et al., 2006). To test the sensitivity of learned structure, each algorithm was run with 500 nonparametric bootstrapping (Friedman et al., 1999). Each directed pair of variables will have a strength value, defined as the proportion of resamples where the edge appeared, and a direction probability, defined as the conditional probability of the specific direction of the edge. The edge would be retained if it passed a stability threshold of 0.90.

Based on the learned structure, we fit parameters of the averaged network by maximum likelihood. Concretely, each node's local model was fitted as a linear regression of that variable on its parents in the final DAG. The five local regressions together specify the joint distribution. Standardized coefficients were obtained by refitting the same DAG on z-scored data. Confidence intervals and significance tests at the outcome node were computed from ordinary least squares regression, which is exact for the local linear-Gaussian model.

*2.4.3 Validation checks*

Two complementary validation checks were performed on the primary network. Predictive accuracy was assessed by 10-fold cross-validation with 5 runs and the mean negative log-likelihood was computed on the held-out fold. For the 10-fold cross-validation with five repetitions: in each fold, the network parameters were refitted on nine-tenths of the data, and the mean negative log-likelihood was recorded. The same procedure was applied to an empty graph that treats all variables as marginally independent as an object of comparison.

Secondly, posterior predictive check was also conducted to assess whether the generated Bayesian model behaves as a faithful generative model. We drew 1000 synthetic observations from the joint distribution implied by the fitted network by ancestral sampling. We then compared simulated and observed marginal means and standard deviations across all five variables, and the simulated and observed marginal histograms of edit distance. Close agreement on both location and spread of every marginal indicates that the network captures not only the conditional dependencies of interest but also the joint shape of the data well enough to function as a generative model.

## 2.5 Analysis 3: Time-resolved regression analysis of fixation-related potentials

The third analysis aimed to address both the timing and strength of hierarchical structure during reading. In previous analyses, we focused on sentence-level eye movements. In this analysis, we examined fixation-related potentials elicited by individual words.

*2.5.1 Include syntactic surprisal to dissociate statistics-based measures*

The regression included five predictors of fixation-related neural activity, so that the hierarchical and statistical accounts could be tested against each other. Syntactic depth is the hierarchical predictor, carried over from Analysis 1 and defined as the number of dependency links from a word to the root of its sentence's dependency parse.

To more precisely dissociate statistics-based measures from syntactic depth, we introduce a new variable, syntactic surprisal. Syntactic surprisal indexed the change in syntactic uncertainty as each word is integrated into an evolving parse structure and captured how well the unfolding sentence matches a predicted syntactic frame. Operationally, syntactic surprisal was estimated from the pretrained joint language-modeling and Combinatory Categorial Grammar supertags LSTM following Arehalli (2022). We loaded the released weights together with the authors' word-to-index and supertag-to-index mappings, and reused their preprocessing pipeline. Each sentence was then passed through the LSTM one token at a time to reproduce the incremental left-to-right processing assumed by surprisal theory. At every step, the model returned a distribution over the next word, a distribution over the supertag of the word just consumed, and an updated hidden state, which together carried the running syntactic context forward to the following token. At the target word, syntactic surprisal was computed by following Equations 3 to 5 of Arehalli (2022). It was derived by marginalizing the prior predicted supertag distribution over the posterior supertag distribution and taking the negative log of the result.

Other variables, including word length, lexical frequency, lexical surprisal, were included as controls. Word frequency was included as it is an important determinant of reading efficiency (Monsell, 1991; Rayner & Raney, 1996; White, 2008; White et al., 2018). High-frequency words are identified more rapidly, receive shorter fixations, and are more likely to be skipped compared to low-frequency words (Inhoff & Rayner, 1986; Rayner & Duffy, 1986). Word surprisal was also included, which was derived from GPT2 as in the previous section. Word length was included as a low-level visual predictor known to influence fixation behavior (Kliegl et al., 2006; McConkie & Rayner, 1975), and produce early modulations of fixation-locked neural activity in co-registered EEG and eye-tracking studies (Dimigen et al., 2011; Hauk et al., 2004).

*2.5.2 FRP extraction and regression*

Fixation-related potentials were extracted by segmenting the continuous EEG signal into epochs time-locked to the onset of each eye fixation, using "EYE EEG extension" (Winkler et al., 2014). To ensure precise temporal alignment between EEG and eye-tracking data, shared events recorded in both data streams were identified, and a linear transformation was applied to correct for minor timing discrepancies between the two modalities. The resulting synchronization error was minimal, not exceeding one sample (2 ms). Following synchronization, the EEG was segmented into epochs ranging from –600 ms to +1000 ms relative to fixation onset.

Regression was conducted separately for all participants, with the five standardized predictors as inputs and the trial-wise EEG epochs as outputs. We used ridge regression to estimate the contribution of each predictor. For each participant, this yields a coefficient time course at each of the 105 channels for each of the five predictors over the –600 to +1000 ms epoch.

*2.5.3 Cluster-based permutation*

Group-level inference on the subject-level regression coefficient time courses was conducted using one-sample, sign-flip cluster-based permutation tests against zero (Maris & Oostenveld, 2007; Sassenhagen & Draschkow, 2019). For each predictor of interest, each subject's channel-level $\beta$ time course was averaged across the channels of each region of interest to yield a single time course per subject per ROI over the –600 to +1000 ms epoch surrounding fixation onset. ROIs were defined a priori from the HydroCel 128-channel sensor net layout by crossing hemisphere (left, right) with anterior–posterior level (frontal, central, posterior). This yielded six topographic ROIs (left frontal, right frontal, left central, right central, left posterior, right posterior). All six ROIs were tested for the predictors.

For each ROI × predictor combination, a one-sample t-test was computed against zero at every time sample. Samples whose t-statistic exceeded the cluster-forming threshold were grouped into contiguous clusters (one-tailed, $\alpha = 0.05$). Each cluster was assigned a mass equal to the sum of the t-statistics it contained. A null distribution of the maximum cluster mass was constructed from 5,000 sign-flip permutations, in which the sign of each subject's full time course was independently flipped before the t-statistics, clusters, and masses were recomputed. Cluster-level p-values were obtained and clusters with $p < .05$ were considered significant. Cluster-permutation testing controls family-wise error at the cluster level within each test. For visualization, group-mean β time courses were smoothed with a Gaussian kernel ($\sigma = 3$), but all statistical inference was carried out on the unsmoothed data.

## 3 Results

### 3.1 Analysis 1: Transitional probability analysis

To test whether eye movements during naturalistic reading are selectively organized around syntactic heads, we compared transitional probabilities between two conditions: the text baseline condition and fixation-derived condition. Transitional probabilities were analyzed separately for four transition types: head-to-head, head-to-non-head, non-head-to-head, and non-head-to-non-head. The prediction was that readers would show a higher probability of transitions between syntactic heads (words with low syntactic depth) in fixation-derived condition in comparison with baseline.

A descriptive analysis of word depth was first visualized in Figure 2c. Across the 400 sentences, depth ranged from 0 to 11 and peaked at depth 2, with a long right tail.

The results showed that readers preferentially moved between low-depth, syntactically central words, and moved less between high-depth, peripheral words, than the linear text order would predict (Figure 2b). Two opposing patterns supported the effect.

For head-to-head transitions, the fixation-derived condition (M = 0.22, SD = 0.20) showed a higher mean probability than the text baseline (M = 0.17, SD = 0.18). A Welch's *t*-test confirmed that this difference was significant ($t = 3.63$, $p = .003$). This pattern suggested that during natural reading, readers were more likely to fixate on and transition between syntactic heads than a baseline. The syntactic heads are also parts of the sentence most informative about syntactic configuration. This indicated that readers tend to use syntactic heads as structural anchors to anticipate how incoming material will fit together. The effect survives when regressions are removed from the fixation sequences ($t = 2.339$, $p = 0.019$). This ruled out an artifact of backward eye movements.

In contrast, non-head-to-non-head transitions fell. The fixation-derived condition (M = 0.47, SD = 0.22) showed lower probabilities than the text baseline (M = 0.51, SD = 0.23; $t = -2.72$, $p = .0067$).

The two other transition types did not differ between conditions. Head-to-non-head transitions were near-identical across the fixation-derived and baseline sequences, as were non-head-to-head transitions. This supported the interpretation that fixation patterns during naturalistic reading tend to drift away from syntactically unimportant words and between syntactically important words.

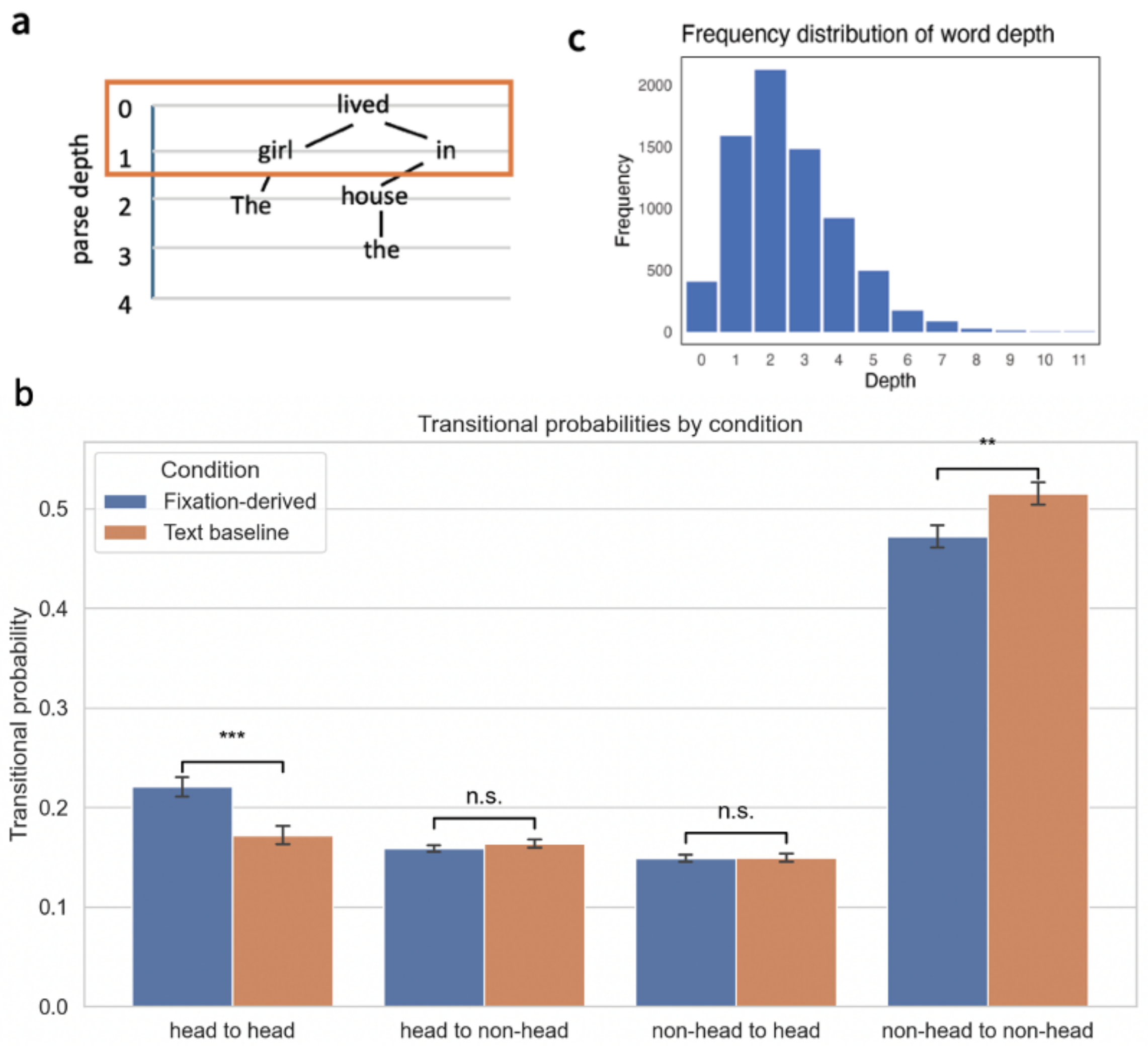

Figure 2. (a) The parsing depths of an example sentence. The words within the orange frame are syntactic heads (with syntactic depth as 0 or 1). (b) Mean transitional probabilities for four transition types across the fixation-derived and text baseline conditions. Error bars indicate ±1 standard error. Asterisks denote statistically significant differences (** $p < .01$; *** $p < .001$). (c) The frequency distribution of word depths across the 400 sentences.

### 3.2 Analysis 2: Bayesian network analysis

This analysis asked whether syntactic depth carried more weight than statistical predictors when both were jointly modeled against scanpath deviation. Scanpath deviation was operationalized as the edit distance between the linear text order and the fixation-derived order. We fit a Gaussian Bayesian network over five variables: maximum syntactic depth, number of clauses, lexical familiarity, surprisal, and edit distance.

### *3.2.1 Correlation*

Edit distance correlated with all four candidate predictors. On a log1p-transformed scale, the Pearson correlations were $r = .69$ with maximum depth, $r = .69$ with familiarity, $r = .59$ with number of clauses, and $r = .44$ with surprisal. Depth and familiarity correlated with edit distance at comparable strength (see Figure 3b for details). This raised the question of whether both contribute directly to scanpath deviation or whether one mediated the effect of the other. We addressed this question with the analysis below.

### *3.2.2 Bootstrap stability and the network skeleton*

Figure 3(c) shows the averaged Bayesian network structure. The learned Bayesian network placed syntactic depth at the root of the structure. Depth projected directly to edit distance and indirectly through familiarity and number of clauses. Familiarity in turn projected to both lexical surprisal and edit distance. Number of clauses was a leaf, connected only to depth; its influence on scanpath deviation was fully mediated through depth rather than acting directly on the edit distance. The edit distance was conditionally dependent on three direct parents: depth, familiarity, and surprisal. Both the hierarchical predictor and the two lexical-statistical predictors thus exerted independent, non-redundant influences on scanpath deviation when admitted to the same model. For other nodes, number of clauses was a leaf, connected only to depth. Its influence on scanpath deviation was fully mediated through depth. This also justified the choice of syntactic depth as the more economical structural variable for indexing the hierarchical organization. The familiarity-to-surprisal edge confirmed that the two lexical-statistical predictors are not independent. Familiarity feeded into the contextual probability estimates that drive surprisal.

The structure was stable across resampling and across algorithms. Across 500 bootstrap resamples, the three direct edges into edit distance and the edge from familiarity to lexical surprisal recovered with direction probability $\geq 0.99$. The edges from depth to number of clauses and from depth to familiarity were directionally indeterminate (direction probabilities 0.48–0.52), but their presence was robust. At the primary stability threshold of 0.90, hill-climbing and tabu search returned identical structures.

### *3.2.3 Effect sizes at the edit distance*

Syntactic depth carried the largest standardized contribution to edit distance among the three direct parents. The local linear-Gaussian regression at edit distance yielded $\beta = 0.40$ for maximum depth, $\beta = 0.37$ for familiarity, and $\beta = 0.19$ for surprisal (Table 1). The three-parent model accounted for 61.4% of the variance in log edit distance ($R^2 = .614$, $F = 209.3$, $p < .001$).

Depth and familiarity contributed at comparable magnitude, with overlapping confidence intervals. Surprisal contributed at half the magnitude of the other two. Maximum depth and number of clauses carried partially overlapping structural information about scanpath organization, with the bulk of the effect captured by depth.

Table 1. Local linear-Gaussian model at edit distance in the primary Bayesian network (threshold = 0.90).

| parent | β (std) | 95% CI | t | *p* |
|---|---|---|---|---|
| Maximum depth | 0.396 | [0.084, 0.126] | 9.71 | <.001 |

| Sentence familiarity | 0.374 | [1.387, 2.167] | 8.96 | <.001 |
|---|---|---|---|---|
| Sentence surprisal | 0.187 | [0.060, 0.125] | 5.56 | <.001 |

*3.2.4 Validation of the primary network*

The primary network outperformed the independence baseline under cross-validation. Mean cross-validated negative log-likelihood was 3.580 (SD = 0.004) for the primary network and 4.667 for the empty graph. The primary structure thus markedly outperformed the independence baseline while using only a small number of arcs.

Posterior predictive checks confirmed that the fitted network reproduces the observed data. Simulated marginals matched observed marginals within sampling noise across all five variables. Means and standard deviations agreed on both location and dispersion (Supplementary Table S2). Histograms of observed and simulated edit distance were visually indistinguishable in shape (Figure 3d).

These analyses placed hierarchy directly against the statistical alternative at the sentence level. Familiarity contributed substantially, but not more than depth; surprisal contributed at roughly half the magnitude of either. Bivariate correlations alone could not have revealed this asymmetry, because depth and familiarity correlate with edit distance at comparable strength. When all three predictors were admitted to the same model, depth retained the largest direct effect on scanpath deviation.

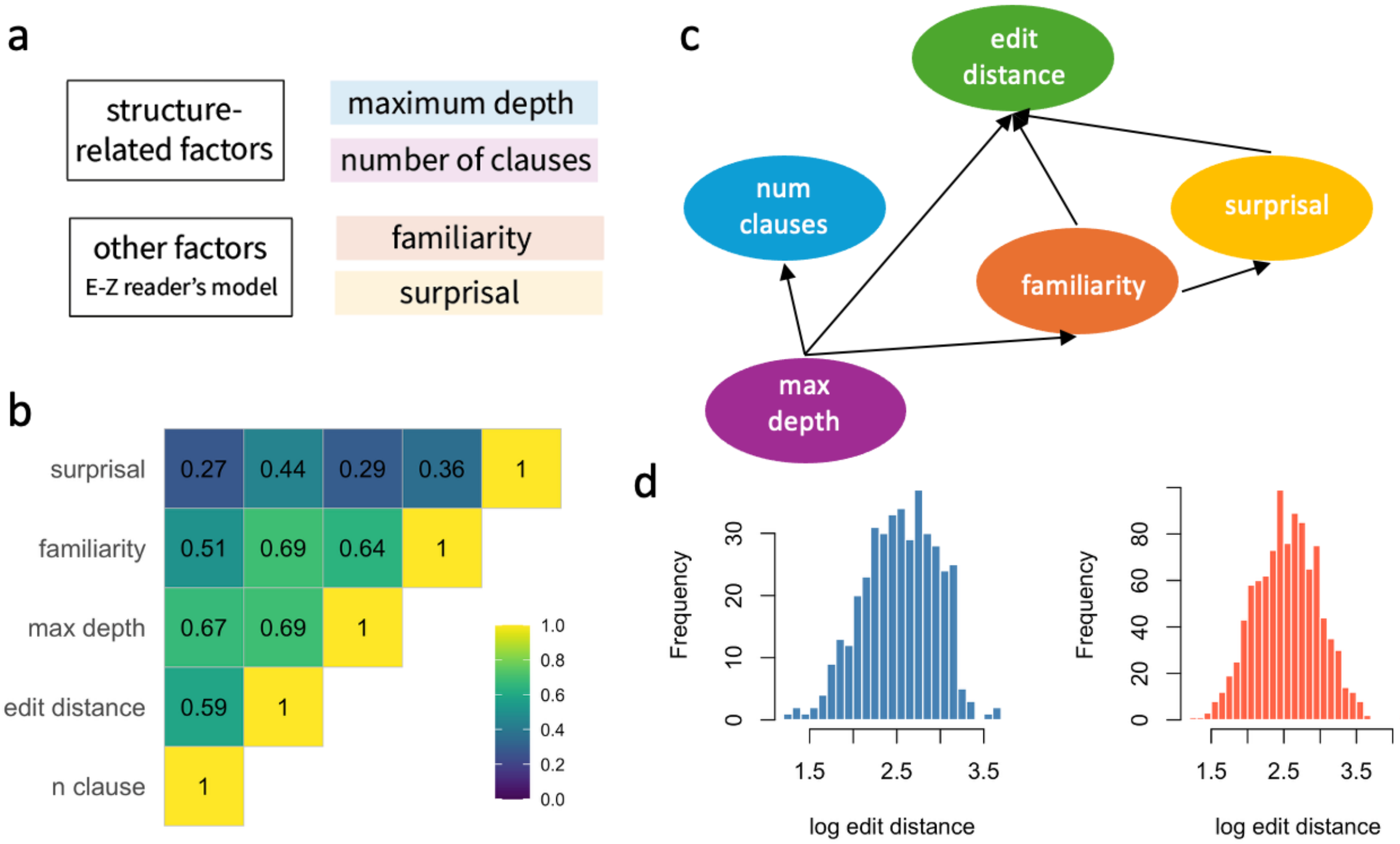

Figure 3. (a) Structural and lexical predictors of scanpath deviation. (b) Correlation matrix showing pairwise associations among structural (number of clauses, maximum depth, edit distance) and lexical (familiarity, surprisal) variables. (c) Averaged Bayesian network at a 0.90 stability threshold, with hill-climbing search across 500 bootstrap resamples. Nodes denote

sentence-level variables; arrows denote directed dependencies retained at the threshold. Maximum depth serves as the root variable, projecting to number of clauses, sentence familiarity, and edit distance. Sentence familiarity in turn projects to surprisal and to edit distance, and sentence surprisal projects to edit distance. Edit distance, the behavioral outcome, thus has three direct parents (depth, familiarity, surprisal) and one indirect contributor mediated by depth (number of clauses). (d) Posterior predictive check for edit distance. Histograms of observed (n = 400, left) and posterior-predictive simulated (n = 1000, right). The similarity between the two supports the adequacy of fitted Bayesian network as a proper generative model.

### 3.3 Analysis 3: Time-resolved regression analysis of fixation-related potentials

Of the five predictors regressed against fixation-related potentials across six ROIs, three produced significant clusters: lexical frequency and lexical surprisal at left-posterior sites, and syntactic depth at right-central sites. Syntactic surprisal and word length produced no significant cluster at any ROI and were thus not visualized (Figure 4).

#### *3.3.1 Comparing the timing of syntactic depth with statistical predictors*

Figure 4 shows the regression coefficient time course for lexical frequency, lexical surprisal, syntactic surprisal and syntactic depth across front left (top panel), mid left (mid panel) and mid right regions (bottom panel). The three significant clusters spanned overlapping time windows around fixation onset. Lexical frequency at left-posterior sites yielded a cluster from –36 to +4 ms ($p$ = .020). Lexical surprisal at left-posterior sites yielded a cluster from –86 to –54 ms ($p$ = .007). Syntactic depth at right-central sites yielded a cluster from –108 to +126 ms ($p$ = .019). The four remaining ROIs (left frontal, right frontal, left central, right posterior) yielded no significant clusters for any predictor.

All three clusters straddled fixation onset. Onsets fell between –130 and –90 ms; offsets fell between +34 and +142 ms. The pre-fixational onset is shared by statistical and structural predictors alike. Lexical frequency, lexical surprisal, and syntactic depth all began to modulate fixation-locked neural activity roughly 90–130 ms before the eyes reach the word, and they continued to modulate it through the earliest stage of foveal processing. The timing was consistent with processing that begins during parafoveal preview and is well in progress by foveal uptake.

The three clusters also differed in scalp topography. The statistical predictors modulate activity over left-posterior region, consistent with the established left-posterior topography of lexical-access and word-level integration effects in the co-registered EEG and eye-tracking literature (Sereno & Rayner, 2003). Syntactic depth modulates activity over right-frontal cortex, dissociated from the lexical effects in both topography and predictor identity. However, as scalp ROIs are not source localizations, this topographic separation should be taken with caution.

reflects coefficient size rather than temporal extent.

#### *3.3.2 Comparing syntactic depth with syntactic surprisal*

Syntactic depth produced a robust cluster; syntactic surprisal produced none. The two variables represent fundamentally different conceptions of syntactic processing. Syntactic surprisal is a probabilistic, statistical measure: it quantifies the unexpectedness of a word's syntactic category and can in principle be computed by a system that tracks probabilities over structural categories without building a hierarchical representation. Syntactic depth, by contrast, is defined only relative to a hierarchical representation. Under a purely statistical account, syntactic surprisal should drive fixation-locked modulation, with depth either non-significant or significant only in proportion to its correlation with surprisal. The opposite pattern was obtained. The variable that required a hierarchical representation produced a robust cluster; the variable that does not require one did not. This dissociation provides positive evidence for hierarchical representations in online reading, because it cannot be reproduced by a model that tracks only flat sequential statistics.

## 4. General Discussion

Whether reading depends on hierarchical syntactic representations or on flatter, statistics-based mechanisms remain debated, and most evidence for hierarchy comes from constructed, lab-based paradigms. We tested the question in naturalistic reading by combining eye movements with electrophysiology, and we asked two things of syntactic depth: when it operates, and how strongly. Across three analyses on co-registered EEG and eye-tracking, the syntactic signal emerged before the eyes reached the word. Readers directed fixations toward syntactically central words before fixating them, and the neural response to syntactic depth arose before fixation onset. Syntactic structure operates anticipatorily, shaping behavior and neural activity before lexical uptake rather than only after a word is encountered.

### 4.1 Syntactic processing in reading is anticipatory

Previous debate on the structure and statistics have generally failed to highlight the temporal profile of structure while controlling structure and statistics at the same time. For studies that do investigate time windows of syntactic processing, three time windows emerged. Firstly, an early time window, that spanning from 100 to 250 ms after an informative cue. Syntactic category information becomes decodable roughly 100 ms after word onset in continuous speech and 160 to 190 ms after within-word disambiguation (Arana et al., 2021; Gwilliams et al., 2024; Matchin et al., 2019). Classical ERP work on word-category violations identified the Early Left Anterior Negativity in roughly the same window, 100 to 200 ms over left-anterior sites, as a signature of first-pass detection that the incoming word does not fit the projected phrase-structural frame (Friederici, 2002; Hahne & Friederici, 1999). In other words, the earliest syntactic window signals a rapid registration of word-category, available before the word has been fully integrated into a sentence-level representation. Secondly, a mid time window (250 ~400 ms). This time window is dominated by reliable structure-engagement effects across left fronto-temporal regions. Paradigms that minimize semantic or temporal confounds (e.g., lists embedded in sentences; simultaneous two-word presentations) consistently report increased activations in left inferior frontal gyrus, left anterior temporal lobe, and posterior temporal lobe around 250 to 400 ms, suggesting active phrase-structural integration in this window (Krogh & Pylkkänen, 2025; Law & Pylkkänen, 2021). Auditory MEG with representational analyses also finds ongoing lexico-syntactic sensitivity in posterior temporal cortex during this period (Matchin et al., 2019; Tyler et al., 2013). These evidence suggests that the mid window is where the phrase-structural computations are focused. Thirdly, a late time window (≥500~900+ ms, extending further) that encompasses reanalysis and repair processes triggered by both category conflicts and phrase-structural demands (Osterhout & Holcomb, 1993). The defining ERP component of this window is the P600, whose amplitude scales with the difficulty of the required structural revision (Kaan et al., 2000b). Sustained left anterior negativities at the position of an expected gap in filler-gap dependencies also extend through this window, and have been interpreted as a signature of holding an active dependency across intervening material until the gap is reached (Phillips et al., 2005b). Across these effects, the late window is the period in which the parser commits to a revised representation and integrates structurally costly material into the unfolding interpretation.

Our study pushes the early window backwards into a genuinely anticipatory phase. Analysis 1 showed that the choice of where to move the eyes was already organized around syntactic heads which indicated that readers possibly approach texts with an anticipatory, pre-existing syntactic tree in mind, thus allocating more attention to words most important to structure-building, i.e. syntactic heads. Analysis 3 put a number on this in the neural signal: the fixation-locked cluster for syntactic depth in right-frontal sites began 108 ms before fixation onset and extends to 126 ms after.

The findings join a large body of work showing that comprehension is predictive and hierarchical structure can play a more important role here. The predictive-coding account holds that the brain projects linguistic expectations ahead of the input: strong context pre-activates the form and category of an upcoming word (DeLong et al., 2005; Lau et al., 2006; Van Berkum et al., 2005). However, in these studies, predictive processing is often cast in probabilistic terms, with surprisal as the quantity the comprehender anticipates (Heilbron et al., 2022). Our anticipation effect is not reducible to that quantity. Syntactic depth can be used to organize reading even after statistical expectation is controlled. Secondly, our findings demonstrate that anticipation is graded and structural: readers biased fixations toward syntactically central words across ordinary text, with no constraint engineered into the stimulus, while prior anticipation has been indexed mostly to specific items, e.g. a particular noun, a particular gap, cued by high-constraint contexts (Altmann & Kamide, 1999; Staub & Clifton Jr, 2006). Thirdly, the The naturalistic predictive-coding account reads prediction from the neural signal during listening, where the rate and order of input are fixed (Heilbron et al., 2022). Reading lets the reader choose where and when to look, and our scanpath shows syntactic depth organizing that choice. Our results thus link the predictive neural signal to overt oculomotor control, joining two literatures, predictive coding in comprehension and eye-movement control in reading, that have largely run in parallel.

### 4.2 Structure also dominates the sentence-level scanpath

How strongly hierarchy drives reading, relative to word statistics, remains contested, and naturalistic tests have returned mixed verdicts. Most of those tests read out processing at the word scale, with mixed results. Some studies showed that hierarchical factors could explain EEG variance above sequential n-gram surprisal during naturalistic listening (Brennan et al., 2016; Brennan & Hale, 2019), while others showed a stronger effect from surprisal (Frank et al., 2015; Frank & Bod, 2011b; Michaelov et al., 2023). Across our own analyses, syntactic depth led its influence in the sentence-level scanpath, though in statistical features are marginally stronger in the word-locked neural response.

At the sentence scale, syntactic depth organized scanpath deviation more strongly than any other property. Depth was the largest direct predictor of edit distance ($\beta = 0.40$), and it stood at the root of the network — the one exogenous variable from which lexical familiarity, number of clauses, and the scanpath itself descended. Surprisal predicted the scanpath reliably but at half depth's magnitude ($\beta = 0.19$). Lexical familiarity matched depth in direct effect ($\beta = 0.37$), yet familiarity was itself a child of depth: deeper sentences drew on less familiar words, so part of what familiarity contributed to the scanpath, depth had already set upstream. Structure therefore did more than predict the scanpath; it organized the variables that predicted it.

A sentence-level outcome might favor any sentence-level predictor, so depth could lead by matching the grain of the measure rather than by capturing hierarchy. The network rules this out. Number of clauses, which is a flat complexity count at the same sentence scale as depth, contributed little to the scanpath once depth was present. Its influence ran through depth at the primary stability threshold, and where it entered as a direct parent its effect was small ($\beta = 0.09$). Depth remained the largest predictor and absorbed the shared structural signal. The advantage is specific to syntactic depth, not to sentence-level granularity.

Recent naturalistic work shows the underestimate is not inevitable once the right controls are in place. Structural features exert broader temporal effects than statistical ones during natural listening (Weissbart & Martin, 2024), and hierarchical processing dominates the word-by-word neural response once memory-based competition among maintained items is modeled (Zacharopoulos et al., 2026). Those studies recover hierarchy's strength at the neural, word-aligned scale; ours recovers it at a behavioral, sentence-level scale they do not address.

The weaker effect of surprisal marks not a weaker role for statistical processing but a different one: surprisal governs the moment-to-moment cost of recognizing the next word, and its purchase on the global shape of the scanpath is correspondingly limited. A division of labor emerges across scales. Syntactic depth organizes the global trajectory of reading; word statistics govern the local cost of each word along it. Thus, although lexical frequency and predictability contribute to local processing differences, their impact on global scan path deviations is more limited. This explains why familiarity had a moderate influence and surprisal a weaker effect, and why surprisal had much lower mutual information with edit distance than parsing depth in the Bayesian model.

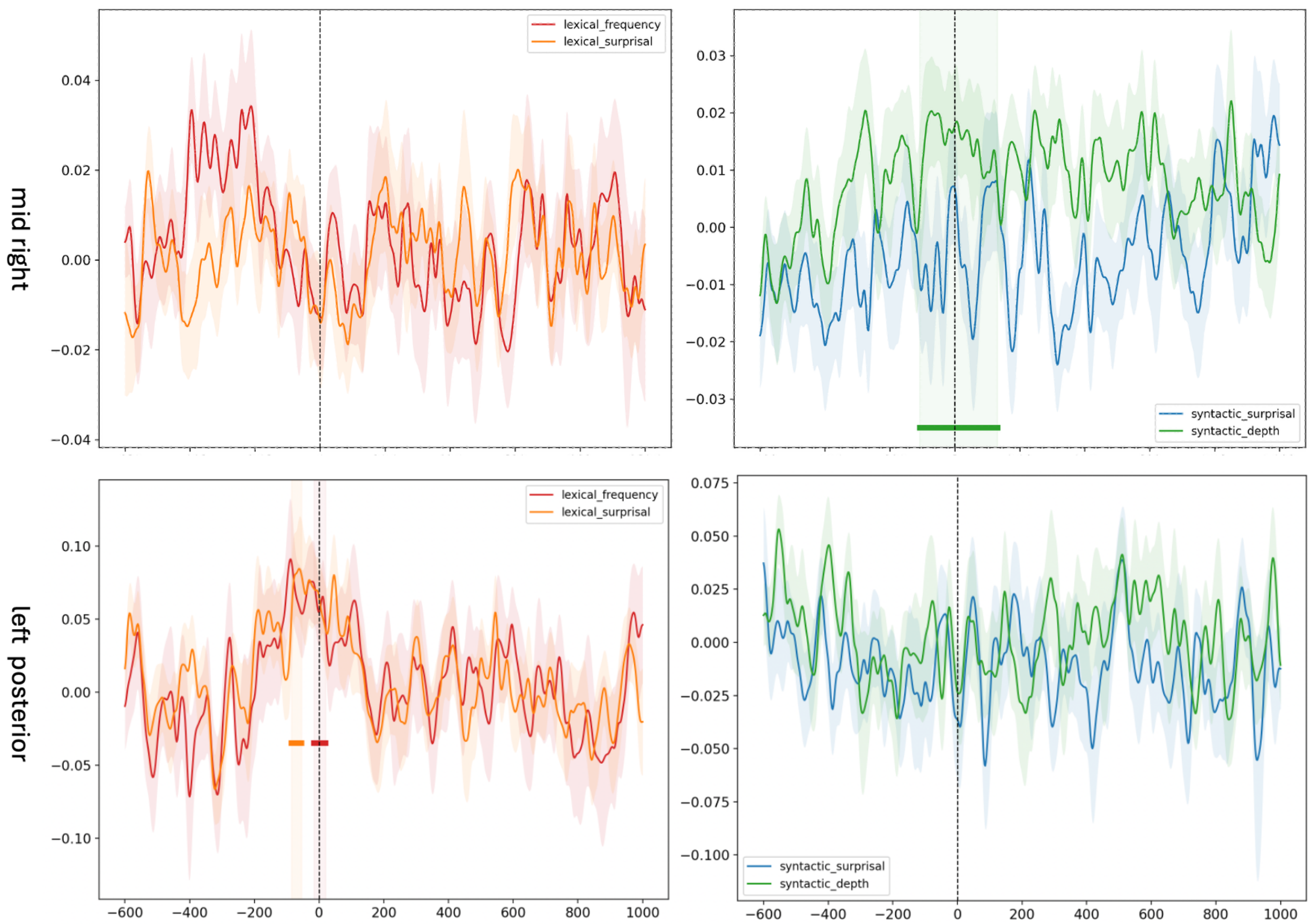

Figure 4. Regression coefficient time course for lexical frequency, lexical surprisal, syntactic surprisal, and syntactic depth in the right-central (top row) and left-posterior (bottom row) regions. The dashed vertical line marks fixation onset of a certain word and the y-axis stands for regression coefficient strength. The shaded regions around each regression line indicate between-subject uncertainty, and the lightly shaded vertical windows indicate time intervals that reached significance in the cluster-based permutation analysis.

## 5. Conclusion

The central debate over whether online comprehension is fundamentally hierarchical or largely reducible to statistical and sequence-based mechanisms has often been run as a contest for a single winner. The present results suggest that structure can play an earlier and more dominant role in reading. Across three analyses on co-registered EEG and eye-tracking from naturalistic reading, hierarchical structure operationalized as syntactic depth left detectable traces on the eye movement path, on the magnitude of departures from linear word-by-word reading, and on word-level fixation-related neural activity.

At the behavioral level, hierarchical depth was the heaviest single contributor to scanpath departures from linear reading in a Bayesian network that contained sentence familiarity and sentence surprisal on equal footing, and the head-to-head transition bias survived after regression-related fixations were removed, ruling out a backward-saccade artifact and a pure linear-adjacency account. At the neural level, syntactic depth tracks word-level activity, and its signature is anticipatory: the fixation-locked cluster at right-frontal sites begins 108 ms before fixation onset and extends to 126 ms after.

These strands of evidence from all three analyses support what an updated account of syntactic processing: readers activate an internalized syntactic structure that (i) privileges head-to-head connections in guiding eye movements, (ii) modulates scanpath primarily through structural depth rather than lexical probability, and (iii) shapes neural activity in both anticipatory and rapid-integration phases. Our findings therefore lend support to hybrid models (Kuperberg, 2007; Staub & Clifton Jr, 2006), in which both syntactic structure and probabilistic expectations jointly shape comprehension, but with syntactic scaffolds providing the backbone of early processing.

**Reference:**

Altmann, G. T. M., & Kamide, Y. (1999). Incremental interpretation at verbs: Restricting the domain of subsequent reference. *Cognition*, *73*(3), 247–264. https://doi.org/10.1016/S0010-0277(99)00059-1

Arana, S., Schoffelen, J.-M., Mitchell, T., & Hagoort, P. (2021). *MVPA does not reveal neural representations of hierarchical linguistic structure in MEG* (p. 2021.02.19.431945). bioRxiv. https://doi.org/10.1101/2021.02.19.431945

Arehalli, S., Dillon, B., & Linzen, T. (2022). Syntactic Surprisal From Neural Models Predicts, But Underestimates, Human Processing Difficulty From Syntactic Ambiguities. In A. Fokkens & V. Srikumar (Eds.), *Proceedings of the 26th Conference on Computational Natural Language Learning (CoNLL)* (pp. 301–313). Association for Computational Linguistics. https://doi.org/10.18653/v1/2022.conll-1.20

Bemis, D. K., & Pylkkänen, L. (2011). Simple Composition: A Magnetoencephalography Investigation into the Comprehension of Minimal Linguistic Phrases. *Journal of Neuroscience*, *31*(8), 2801–2814. https://doi.org/10.1523/JNEUROSCI.5003-10.2011

Brennan, J. R., & Hale, J. T. (2019). Hierarchical structure guides rapid linguistic predictions during naturalistic listening. *PLOS ONE*, *14*(1), e0207741. https://doi.org/10.1371/journal.pone.0207741

Brennan, J. R., Stabler, E. P., Van Wagenen, S. E., Luh, W.-M., & Hale, J. T. (2016). Abstract linguistic structure correlates with temporal activity during naturalistic comprehension. *Brain and Language*, *157–158*, 81–94. https://doi.org/10.1016/j.bandl.2016.04.008

Caucheteux, C., Gramfort, A., & King, J.-R. (2023). Evidence of a predictive coding hierarchy in the human brain listening to speech. *Nature Human Behaviour*, *7*(3), 430–441. https://doi.org/10.1038/s41562-022-01516-2

Christiansen, M. H., & Chater, N. (2008). Language as shaped by the brain. *The Behavioral and Brain Sciences*, *31*(5), 489–508; discussion 509-558. https://doi.org/10.1017/S0140525X08004998

de Marneffe, M.-C., MacCartney, B., & Manning, C. D. (2006). Generating Typed Dependency Parses from Phrase Structure Parses. In N. Calzolari, K. Choukri, A. Gangemi, B. Maegaard, J. Mariani, J. Odijk, & D. Tapias (Eds.), *Proceedings of the Fifth International Conference on Language Resources and Evaluation (LREC'06)*. European Language Resources Association (ELRA). https://aclanthology.org/L06-1260/

DeLong, K. A., Urbach, T. P., & Kutas, M. (2005). Probabilistic word pre-activation during language comprehension inferred from electrical brain activity. *Nature Neuroscience*, *8*(8), 1117–1121. https://doi.org/10.1038/nn1504

Dempsey, J., Tsiola, A., & Christianson, K. (2023). Eye-tracking evidence from attachment structures favors a serial model of discourse–sentence interactivity. *Discourse Processes*. (world). https://www.tandfonline.com/doi/abs/10.1080/0163853X.2023.2260246

Dillon, B., Mishler, A., Sloggett, S., & Phillips, C. (2013). Contrasting intrusion profiles for agreement and anaphora: Experimental and modeling evidence. *Journal of Memory and Language*, *69*(2), 85–103. https://doi.org/10.1016/j.jml.2013.04.003

Dimigen, O., Sommer, W., Hohlfeld, A., Jacobs, A. M., & Kliegl, R. (2011). Coregistration of eye movements and EEG in natural reading: Analyses and review. *Journal of Experimental Psychology. General*, *140*(4), 552–572. https://doi.org/10.1037/a0023885

Ding, N., Melloni, L., Zhang, H., Tian, X., & Poeppel, D. (2016). Cortical tracking of hierarchical linguistic structures in connected speech. *Nature Neuroscience*, *19*(1), 158–164. https://doi.org/10.1038/nn.4186

Frank, S. L., & Bod, R. (2011a). Insensitivity of the human sentence-processing system to hierarchical structure. *Psychological Science*, *22*(6), 829–834. https://doi.org/10.1177/0956797611409589

Frank, S. L., & Bod, R. (2011b). Insensitivity of the human sentence-processing system to hierarchical structure. *Psychological Science*, *22*(6), 829–834. https://doi.org/10.1177/0956797611409589

Frank, S. L., Bod, R., & Christiansen, M. H. (2012). How hierarchical is language use? *Proceedings of the Royal Society B: Biological Sciences*, *279*(1747), 4522–4531. https://doi.org/10.1098/rspb.2012.1741

Frank, S. L., Otten, L. J., Galli, G., & Vigliocco, G. (2015). The ERP response to the amount of information conveyed by words in sentences. *Brain and Language*, *140*, 1–11. https://doi.org/10.1016/j.bandl.2014.10.006

Frazier, L., & Rayner, K. (1982). Making and correcting errors during sentence comprehension: Eye movements in the analysis of structurally ambiguous sentences. *Cognitive Psychology*, *14*(2), 178–210. https://doi.org/10.1016/0010-0285(82)90008-1

Friederici, A. D. (2002). Towards a neural basis of auditory sentence processing. *Trends in Cognitive Sciences*, *6*(2), 78–84.

Friedman, N., Goldszmidt, M., & Wyner, A. (1999). *On the Application of The Bootstrap for Computing Confidence Measures on Features of Induced Bayesian Networks*.

Glover, F., & Marti, R. (2006). Tabu Search. In E. Alba & R. Martí (Eds.), *Metaheuristic Procedures for Training Neutral Networks* (pp. 53–69). Springer US. https://doi.org/10.1007/0-387-33416-5_3

Goodkind, A., & Bicknell, K. (2018). Predictive power of word surprisal for reading times is a linear function of language model quality. In A. Sayeed, C. Jacobs, T. Linzen, & M. van Schijndel (Eds.), *Proceedings of the 8th Workshop on Cognitive Modeling and Computational Linguistics (CMCL 2018)* (pp. 10–18). Association for Computational Linguistics. https://doi.org/10.18653/v1/W18-0102

Gwilliams, L., Marantz, A., Poeppel, D., & King, J.-R. (2024). Top-down information shapes lexical processing when listening to continuous speech. *Language, Cognition and Neuroscience*. (world). https://www.tandfonline.com/doi/abs/10.1080/23273798.2023.2171072

Gwilliams, L., Marantz, A., Poeppel, D., & King, J.-R. (2025). Hierarchical dynamic coding coordinates speech comprehension in the human brain. *Proceedings of the National Academy of Sciences*, *122*(42), e2422097122. https://doi.org/10.1073/pnas.2422097122

Hahne, A., & Friederici, A. D. (1999). Electrophysiological evidence for two steps in syntactic analysis. Early automatic and late controlled processes. *Journal of Cognitive Neuroscience*, *11*(2), 194–205. https://doi.org/10.1162/089892999563328

Hale, J. (2001). A Probabilistic Earley Parser as a Psycholinguistic Model. *Second Meeting of the North American Chapter of the Association for Computational Linguistics*. NAACL 2001. https://aclanthology.org/N01-1021/

Hauk, O., Johnsrude, I., & Pulvermüller, F. (2004). Somatotopic representation of action words in human motor and premotor cortex. *Neuron*, *41*(2), 301–307. https://doi.org/10.1016/s0896-6273(03)00838-9

Heckerman, D., Geiger, D., & Chickering, D. M. (1995). Learning Bayesian Networks: The Combination of Knowledge and Statistical Data. *Machine Learning*, *20*(3), 197–243. https://doi.org/10.1023/A:1022623210503

Heilbron, M., Armeni, K., Schoffelen, J.-M., Hagoort, P., & de Lange, F. P. (2022). A hierarchy of linguistic predictions during natural language comprehension. *Proceedings of the National Academy of Sciences*, *119*(32), e2201968119. https://doi.org/10.1073/pnas.2201968119

Hollenstein, N., Rotsztejn, J., Troendle, M., Pedroni, A., Zhang, C., & Langer, N. (2018). ZuCo, a simultaneous EEG and eye-tracking resource for natural sentence reading. *Scientific Data*, *5*(1), 180291. https://doi.org/10.1038/sdata.2018.291

Huang, K.-J., Arehalli, S., Kugemoto, M., Muxica, C., Prasad, G., Dillon, B., & Linzen, T. (2024). Large-scale benchmark yields no evidence that language model surprisal explains syntactic disambiguation difficulty. *Journal of Memory and Language*, *137*, 104510. https://doi.org/10.1016/j.jml.2024.104510

Hultén, A., Schoffelen, J.-M., Uddén, J., Lam, N. H. L., & Hagoort, P. (2019). How the brain makes sense beyond the processing of single words—An MEG study. *NeuroImage*, *186*, 586–594. https://doi.org/10.1016/j.neuroimage.2018.11.035

Inhoff, A. W., & Rayner, K. (1986). Parafoveal word processing during eye fixations in reading: Effects of word frequency. *Perception & Psychophysics*, *40*(6), 431–439. https://doi.org/10.3758/BF03208203

Jurafsky, D. (1996). A Probabilistic Model of Lexical and Syntactic Access and Disambiguation. *Cognitive Science*, *20*(2), 137–194. https://doi.org/10.1207/s15516709cog2002_1

Kaan, E., Harris, A., Gibson, E., & Holcomb, P. (2000a). The P600 as an index of syntactic integration difficulty. *Language and Cognitive Processes*, *15*(2), 159–201. https://doi.org/10.1080/016909600386084

Kaan, E., Harris, A., Gibson, E., & Holcomb, P. (2000b). The P600 as an index of syntactic integration difficulty. *Language and Cognitive Processes*, *15*(2), 159–201. https://doi.org/10.1080/016909600386084

Kliegl, R., Nuthmann, A., & Engbert, R. (2006). Tracking the mind during reading: The influence of past, present, and future words on fixation durations. *Journal of Experimental Psychology: General*, *135*(1), 12–35. https://doi.org/10.1037/0096-3445.135.1.12

Krogh, S., & Pylkkänen, L. (2025). *Manipulating syntax without taxing working memory: MEG correlates of syntactic dependencies in a Verb-Second language* (p. 2024.02.20.581245). bioRxiv. https://doi.org/10.1101/2024.02.20.581245

Kuperberg, G. R. (2007). Neural mechanisms of language comprehension: Challenges to syntax. *Brain Research*, *1146*, 23–49.

Kutas, M., & Federmeier, K. D. (2011). Thirty Years and Counting: Finding Meaning in the N400 Component of the Event-Related Brain Potential (ERP). *Annual Review of Psychology*, *62*(1), 621–647. https://doi.org/10.1146/annurev.psych.093008.131123

Lau, E., Stroud, C., Plesch, S., & Phillips, C. (2006). The role of structural prediction in rapid syntactic analysis. *Brain and Language*, *98*(1), 74–88.

Law, R., & Pylkkänen, L. (2021). Lists with and without Syntax: A New Approach to Measuring the Neural Processing of Syntax. *Journal of Neuroscience*, *41*(10), 2186–2196. https://doi.org/10.1523/JNEUROSCI.1179-20.2021

Leiken, K., McElree, B., & Pylkkänen, L. (2015). Filling Predictable and Unpredictable Gaps, with and without Similarity-Based Interference: Evidence for LIFG Effects of Dependency Processing. *Frontiers in Psychology*, *6*. https://doi.org/10.3389/fpsyg.2015.01739

Levy, R. (2008). Expectation-based syntactic comprehension. *Cognition*, *106*(3), 1126–1177. https://doi.org/10.1016/j.cognition.2007.05.006

Li, J., & Pylkkänen, L. (2021). Disentangling Semantic Composition and Semantic Association in the Left Temporal Lobe. *Journal of Neuroscience*, *41*(30), 6526–6538. https://doi.org/10.1523/JNEUROSCI.2317-20.2021

Liu, W., Xiang, M., & Ding, N. (2026). Active use of latent tree-structured sentence representation in humans and large language models. *Nature Human Behaviour*, *10*(2), 303–316. https://doi.org/10.1038/s41562-025-02297-0

Lu, X. (2010). Automatic analysis of syntactic complexity in second language writing. *International Journal of Corpus Linguistics*, *15*(4), 474–496. https://doi.org/10.1075/ijcl.15.4.02lu

Malsburg, T. von der, & Vasishth, S. (2013). Scanpaths reveal syntactic underspecification and reanalysis strategies. *Language and Cognitive Processes*. (world). https://www.tandfonline.com/doi/abs/10.1080/01690965.2012.728232

Maris, E., & Oostenveld, R. (2007). Nonparametric statistical testing of EEG- and MEG-data. *Journal of Neuroscience Methods*, *164*(1), 177–190. https://doi.org/10.1016/j.jneumeth.2007.03.024

Matchin, W., Brodbeck, C., Hammerly, C., & Lau, E. (2019). The temporal dynamics of structure and content in sentence comprehension: Evidence from fMRI-constrained MEG. *Human Brain Mapping*, *40*(2), 663–678. https://doi.org/10.1002/hbm.24403

McConkie, G. W., & Rayner, K. (1975). The span of the effective stimulus during a fixation in reading. *Perception & Psychophysics*, *17*(6), 578–586. https://doi.org/10.3758/BF03203972

Michaelov, J. A., Arnett, C., Chang, T. A., & Bergen, B. K. (2023). Structural Priming Demonstrates Abstract Grammatical Representations in Multilingual Language Models. In H. Bouamor, J. Pino, & K. Bali (Eds.), *Proceedings of the 2023 Conference on*

*Empirical Methods in Natural Language Processing* (pp. 3703–3720). Association for Computational Linguistics. https://doi.org/10.18653/v1/2023.emnlp-main.227

Monsell, S. (1991). The Nature And Locus Of Word Frequency Effects In Reading. In *Basic Processes in Reading*. Routledge.

Needleman, S. B., & Wunsch, C. D. (1970). A general method applicable to the search for similarities in the amino acid sequence of two proteins. *Journal of Molecular Biology*, *48*(3), 443–453. https://doi.org/10.1016/0022-2836(70)90057-4

Nelson, M. J., El Karoui, I., Giber, K., Yang, X., Cohen, L., Koopman, H., Cash, S. S., Naccache, L., Hale, J. T., Pallier, C., & Dehaene, S. (2017). Neurophysiological dynamics of phrase-structure building during sentence processing. *Proceedings of the National Academy of Sciences*, *114*(18), E3669–E3678. (world). https://doi.org/10.1073/pnas.1701590114

Newell, A. (with Carnegie-Mellon University Computer Science Department). (1973). *Productions systems: Models of control structures*. Carnegie-Mellon University, Department of Computer Science.

Nivre, J., de Marneffe, M.-C., Ginter, F., Goldberg, Y., Hajič, J., Manning, C. D., McDonald, R., Petrov, S., Pyysalo, S., Silveira, N., Tsarfaty, R., & Zeman, D. (2016). Universal Dependencies v1: A Multilingual Treebank Collection. In N. Calzolari, K. Choukri, T. Declerck, S. Goggi, M. Grobelnik, B. Maegaard, J. Mariani, H. Mazo, A. Moreno, J. Odijk, & S. Piperidis (Eds.), *Proceedings of the Tenth International Conference on Language Resources and Evaluation (LREC'16)* (pp. 1659–1666). European Language Resources Association (ELRA). https://aclanthology.org/L16-1262/

Osterhout, L., & Holcomb, P. J. (1993). Event-related potentials and syntactic anomaly: Evidence of anomaly detection during the perception of continuous speech. *Language and Cognitive Processes*. (world). https://doi.org/10.1080/01690969308407584

Phillips, C., Kazanina, N., & Abada, S. H. (2005a). ERP effects of the processing of syntactic long-distance dependencies. *Cognitive Brain Research*, *22*(3), 407–428. https://doi.org/10.1016/j.cogbrainres.2004.09.012

Phillips, C., Kazanina, N., & Abada, S. H. (2005b). ERP effects of the processing of syntactic long-distance dependencies. *Brain Research. Cognitive Brain Research*, *22*(3), 407–428. https://doi.org/10.1016/j.cogbrainres.2004.09.012

Price, A. R., Bonner, M. F., Peelle, J. E., & Grossman, M. (2015). Converging Evidence for the Neuroanatomic Basis of Combinatorial Semantics in the Angular Gyrus. *Journal of Neuroscience*, *35*(7), 3276–3284. https://doi.org/10.1523/JNEUROSCI.3446-14.2015

Radford, A., Wu, J., Child, R., Luan, D., Amodei, D., & Sutskever, I. (2019). *Language Models are Unsupervised Multitask Learners*.

Rayner, K., & Duffy, S. A. (1986). Lexical complexity and fixation times in reading: Effects of word frequency, verb complexity, and lexical ambiguity. *Memory & Cognition*, *14*(3), 191–201. https://doi.org/10.3758/BF03197692

Rayner, K., & Raney, G. E. (1996). Eye movement control in reading and visual search: Effects of word frequency. *Psychonomic Bulletin & Review*, *3*(2), 245–248. https://doi.org/10.3758/BF03212426

Reichle, E. D., Pollatsek, A., Fisher, D. L., & Rayner, K. (1998). Toward a model of eye movement control in reading. *Psychological Review*, *105*(1), 125–157. https://doi.org/10.1037/0033-295x.105.1.125

Reichle, E. D., Rayner, K., & Pollatsek, A. (2003). The E-Z reader model of eye-movement control in reading: Comparisons to other models. *The Behavioral and Brain Sciences*, *26*(4), 445–476; discussion 477-526. https://doi.org/10.1017/s0140525x03000104

Sassenhagen, J., & Draschkow, D. (2019). Cluster-based permutation tests of MEG/EEG data do not establish significance of effect latency or location. *Psychophysiology*, *56*(6), e13335. https://doi.org/10.1111/psyp.13335

Schrimpf, M., Blank, I. A., Tuckute, G., Kauf, C., Hosseini, E. A., Kanwisher, N., Tenenbaum, J. B., & Fedorenko, E. (2021). The neural architecture of language: Integrative modeling converges on predictive processing. *Proceedings of the National Academy of Sciences of the United States of America*, *118*(45), e2105646118. https://doi.org/10.1073/pnas.2105646118

Schuster, S., & Manning, C. D. (2016). Enhanced English Universal Dependencies: An Improved Representation for Natural Language Understanding Tasks. In N. Calzolari, K. Choukri, T. Declerck, S. Goggi, M. Grobelnik, B. Maegaard, J. Mariani, H. Mazo, A. Moreno, J. Odijk, & S. Piperidis (Eds.), *Proceedings of the Tenth International Conference on Language Resources and Evaluation (LREC'16)* (pp. 2371–2378). European Language Resources Association (ELRA). https://aclanthology.org/L16-1376/

Scutari, M. (2010). Learning Bayesian Networks with the bnlearn R Package. *Journal of Statistical Software*, *35*, 1–22. https://doi.org/10.18637/jss.v035.i03

Sereno, S. C., & Rayner, K. (2003). Measuring word recognition in reading: Eye movements and event-related potentials. *Trends in Cognitive Sciences*, *7*(11), 489–493. https://doi.org/10.1016/j.tics.2003.09.010

Shain, C., Meister, C., Pimentel, T., Cotterell, R., & Levy, R. (2024). Large-scale evidence for logarithmic effects of word predictability on reading time. *Proceedings of the National Academy of Sciences*, *121*(10), e2307876121. https://doi.org/10.1073/pnas.2307876121

Socher, R., Perelygin, A., Wu, J., Chuang, J., Manning, C. D., Ng, A., & Potts, C. (2013). Recursive Deep Models for Semantic Compositionality Over a Sentiment Treebank. In D. Yarowsky, T. Baldwin, A. Korhonen, K. Livescu, & S. Bethard (Eds.), *Proceedings of the 2013 Conference on Empirical Methods in Natural Language Processing* (pp. 1631–1642). Association for Computational Linguistics. https://aclanthology.org/D13-1170/

Staub, A., & Clifton Jr, C. (2006). Syntactic prediction in language comprehension: Evidence from either... or. *Journal of Experimental Psychology: Learning, Memory, and Cognition*, *32*(2), 425.

Stowe, L. A. (1985). Parsing WH-constructions: Evidence for on-line gap location. *Language and Cognitive Processes*, *1*(3), 227–245. https://doi.org/10.1080/01690968608407062

Stromswold, K., Caplan, D., Alpert, N., & Rauch, S. (1996). Localization of syntactic comprehension by positron emission tomography. *Brain and Language*, *52*(3), 452–473. https://doi.org/10.1006/brln.1996.0024

Tsamardinos,  Ioannis, Brown, L. E., & Aliferis, C. F. (2006). The Max-Min Hill-Climbing Bayesian network structure learning algorithm. *Machine Learning*, *65*(1), 31–78. https://doi.org/10.1007/s10994-006-6889-7

Tuckute, G., Hansen, S. T., Kjaer, T. W., & Hansen, L. K. (2021). Real-Time Decoding of Attentional States Using Closed-Loop EEG Neurofeedback. *Neural Computation*, *33*(4), 967–1004. https://doi.org/10.1162/neco_a_01363

Tyler, L. K., Cheung, T. P., Devereux, B. J., & Clarke, A. (2013). Syntactic Computations in the Language Network: Characterizing Dynamic Network Properties Using Representational Similarity Analysis. *Frontiers in Psychology*, *4*. https://doi.org/10.3389/fpsyg.2013.00271

Van Berkum, J. J. A., Brown, C. M., Zwitserlood, P., Kooijman, V., & Hagoort, P. (2005). Anticipating Upcoming Words in Discourse: Evidence From ERPs and Reading Times. *Journal of Experimental Psychology: Learning, Memory, and Cognition*, *31*(3), 443–467. https://doi.org/10.1037/0278-7393.31.3.443

van Schijndel, M., & Linzen, T. (2019). Can Entropy Explain Successor Surprisal Effects in Reading? In G. Jarosz, M. Nelson, B. O'Connor, & J. Pater (Eds.), *Proceedings of the Society for Computation in Linguistics (SCiL) 2019* (pp. 1–7). https://doi.org/10.7275/qtbb-9d05

Vosse, T., & Kempen, G. (2000). Syntactic structure assembly in human parsing: A computational model based on competitive inhibition and a lexicalist grammar. *Cognition*, *75*(2), 105–143. https://doi.org/10.1016/S0010-0277(00)00063-9

Wagers, M. W., Lau, E. F., & Phillips, C. (2009). Agreement attraction in comprehension: Representations and processes. *Journal of Memory and Language*, *61*(2), 206–237. https://doi.org/10.1016/j.jml.2009.04.002

Weissbart, H., & Martin, A. E. (2024). The structure and statistics of language jointly shape cross-frequency neural dynamics during spoken language comprehension. *Nature Communications*, *15*(1), 8850. https://doi.org/10.1038/s41467-024-53128-1

Westerlund, M., & Pylkkänen, L. (2014). The role of the left anterior temporal lobe in semantic composition vs. Semantic memory. *Neuropsychologia*, *57*, 59–70. https://doi.org/10.1016/j.neuropsychologia.2014.03.001

White, S. J. (2008). Eye movement control during reading: Effects of word frequency and orthographic familiarity. *Journal of Experimental Psychology: Human Perception and Performance*, *34*(1), 205–223. https://doi.org/10.1037/0096-1523.34.1.205

White, S. J., Drieghe, D., Liversedge, S. P., & Staub, A. (2018). The word frequency effect during sentence reading: A linear or nonlinear effect of log frequency? *Quarterly Journal of Experimental Psychology*, *71*(1), 46–55. https://doi.org/10.1080/17470218.2016.1240813

Williams, A., Reddigari, S., & Pylkkänen, L. (2017). Early sensitivity of left perisylvian cortex to relationality in nouns and verbs. *Neuropsychologia*, *100*, 131–143. https://doi.org/10.1016/j.neuropsychologia.2017.04.029

Winkler, I., Brandl, S., Horn, F., Waldburger, E., Allefeld, C., & Tangermann, M. (2014). Robust artifactual independent component classification for BCI practitioners. *Journal of Neural Engineering*, *11*(3), 035013. https://doi.org/10.1088/1741-2560/11/3/035013

Winkler, I., Haufe, S., & Tangermann, M. (2011). Automatic Classification of Artifactual ICA-Components for Artifact Removal in EEG Signals. *Behavioral and Brain Functions*, *7*(1), 30. https://doi.org/10.1186/1744-9081-7-30

Zacharopoulos, C.-N., Dehaene, S., & Lakretz, Y. (2026). Disentangling Hierarchical and Sequential Computations during Sentence Processing. *Cortex*. https://doi.org/10.1016/j.cortex.2026.02.004

Zhang, L., & Pylkkänen, L. (2015). The interplay of composition and concept specificity in the left anterior temporal lobe: An MEG study. *NeuroImage*, *111*, 228–240. https://doi.org/10.1016/j.neuroimage.2015.02.028

Zou, J., Poeppel, D., & Ding, N. (2026). Constituent-constrained word prediction during language comprehension. *Nature Neuroscience*, 1–12. https://doi.org/10.1038/s41593-026-02272-6